\documentclass[letterpaper, 10 pt, conference]{ieeeconf}  

\IEEEoverridecommandlockouts                              

\usepackage{graphics} 
\usepackage{graphicx}
\usepackage{epsfig} 
\usepackage{times} 
\usepackage{amsmath} 
\usepackage{amssymb}  
\usepackage[english]{babel}
\usepackage{amsthm}

\DeclareMathOperator*{\argmax}{arg\,max}
\DeclareMathOperator*{\argmin}{arg\,min}
\usepackage{bm}
\usepackage[linesnumbered,ruled,vlined]{algorithm2e}
\usepackage[noend]{algpseudocode}
\usepackage[font={small}]{caption}
\usepackage{subcaption}
\usepackage{physics}
\usepackage{xcolor}
\usepackage{tikz}

\usetikzlibrary {arrows.meta}
\usetikzlibrary{fit,
                positioning}
\usepackage[belowskip=1ex]{subcaption}  

\usetikzlibrary{calc} 
\usetikzlibrary{positioning}
\usepackage{balance}
\usepackage{soul}
\usetikzlibrary{arrows}
\usepackage{cite}
\usepackage{url}
\usepackage{multirow}
\usepackage{booktabs} 
\usepackage[hidelinks]{hyperref}

\usepackage{amssymb}
\usepackage{amsfonts}
\usepackage{amsmath}
\usepackage{amsthm}
\usepackage{bm}

\usepackage[normalem]{ulem}                                        
\usepackage{marginnote}
\usepackage[textwidth=10ex,colorinlistoftodos]{todonotes}

\colorlet{mh}{red}
\colorlet{fwu}{red}
\colorlet{ywu}{blue}
\colorlet{kchen}{blue}
\colorlet{lchen}{green}
\colorlet{zbing}{green}
\colorlet{shaddadin}{purple}
\colorlet{iperez}{cyan}
\colorlet{schneider}{magenta}

\usepackage{amssymb}
\usepackage{mathtools}
\usepackage{sansmath}

\mathchardef\mhyphen="2D   

\newcommand{\RNum}[1]{\uppercase\expandafter{\romannumeral #1\relax}}

\usepackage{setspace}
\title{\LARGE \bf

Deep Fuzzy Optimization for Batch-Size and Nearest \\Neighbors in Optimal Robot Motion Planning
}

\author{Liding Zhang$^{1}$, Qiyang Zong$^{1}$, Yu Zhang$^{1}$, Zhenshan Bing$^{1,2}$,  Alois Knoll$^{1\dagger}$ 
\thanks{$^{1}$L. Zhang, Q. Zong, Y, Zhang, Z. Bing, and A. Knoll are with the School of Computation, Information and Technology (CIT), Technical University of Munich, 80333 Munich, Germany.
{\tt\small liding.zhang@tum.de}}%
\thanks{$^{2}$Z. Bing is also with the State Key Laboratory for Novel Software Technology and the School of Science and Technology, Nanjing University (Suzhou Campus), China.~\textit{(Corresponding author: Zhenshan Bing.)}}%
\thanks{$^{\dagger}$The authors acknowledge the financial support by the Bavarian State Ministry for Economic Affairs, Regional Development and Energy (StMWi) for the Lighthouse Initiative KI.FABRIK (Phase 1: Infrastructure and the research and development program under grant no. DIK0249).}
}
\begin{document}

\maketitle
\thispagestyle{empty}
\pagestyle{empty}

\begin{abstract}
Efficient motion planning algorithms are essential in robotics. Optimizing essential parameters, such as batch size and nearest neighbor selection in sampling-based methods, can enhance performance in the planning process. However, existing approaches often lack environmental adaptability. Inspired by the method of the deep fuzzy neural networks, this work introduces Learning-based Informed Trees (LIT*), a sampling-based deep fuzzy learning-based planner that dynamically adjusts batch size and nearest neighbor parameters to obstacle distributions in the configuration spaces. By encoding both global and local ratios via valid and invalid states, LIT* differentiates between obstacle-sparse and obstacle-dense regions, leading to lower-cost paths and reduced computation time. Experimental results in high-dimensional spaces demonstrate that LIT* achieves faster convergence and improved solution quality. It outperforms state-of-the-art single-query, sampling-based planners in environments ranging from $\mathbb{R}^8$ to $\mathbb{R}^{14}$ and is successfully validated on a dual-arm robot manipulation task.
A video showcasing our experimental results is available at: \href{https://youtu.be/NrNs9zebWWk}{\textcolor{blue}{https://youtu.be/NrNs9zebWWk}}.



%
%
%
\end{abstract}

\section{Introduction}
Motion planning is a fundamental problem in robotics and autonomous systems, particularly in high-dimensional continuous spaces~\cite{zhang2024review}. Traditional graph-based algorithms, such as Dijkstra’s algorithm~\cite{dijkstra1959note} and A*~\cite{hart1968formal}, rely on discretized search spaces, making them computationally expensive and resolution-dependent. In contrast, sampling-based methods, such as Rapidly-exploring Random Trees (RRT)~\cite{LaValle1998RRT} and Probabilistic Roadmaps (PRM)~\cite{Kavraki1996PRM}, achieve more efficient exploration. The improved variants, PRM* and RRT*, further guarantee asymptotic optimality~\cite{Karaman2011RRTstar}. Informed RRT*~\cite{gammell2014informed} refines RRT* by restricting sampling to an ellipsoid derived from the initial solution cost, accelerating convergence.

Several algorithms have extended these approaches. Fast Marching Tree (FMT*)~\cite{fmt2015} leverages batch processing for computational efficiency. Batch Informed Trees (BIT*)~\cite{gammell2020batch} balances exploration and optimization through incremental batch processing. Effort Informed Trees (EIT*)~\cite{strub2022adaptively} prioritizes regions with higher potential for path improvement, while Adaptively Informed Trees (AIT*)~\cite{strub2022adaptively} refines computational efficiency through heuristics. These methods have advanced sampling-based motion planning~\cite{zhang25ral}, making them widely applicable in navigation and bio-robotics.

To further improve planning efficiency, researchers have explored learning-based approaches~\cite{ZHANG2025git}. Motion Planning Networks (MPNet)~\cite{Qureshi2020MPNet} eliminate explicit trajectory optimization by directly mapping sensory inputs to motion sequences. Neural Exploration-Exploitation Trees (NEXT)~\cite{chen2020learningplanhighdimensions} and Neural Informed RRT*~\cite{NeuralInformedRRT} integrate neural networks to guide sampling, reducing unnecessary exploration. Deep fuzzy methodologies have also been introduced to enhance adaptability. Wu et al.~\cite{DeepFuzzyRobotSpeed} proposed a deep fuzzy framework that refines velocity commands through defuzzification.
\begin{figure}[t!]
    \centering
    \begin{tikzpicture}

    \node[inner sep=0pt] (russell) at (-2.5,0)
    {\includegraphics[width=0.48\textwidth]{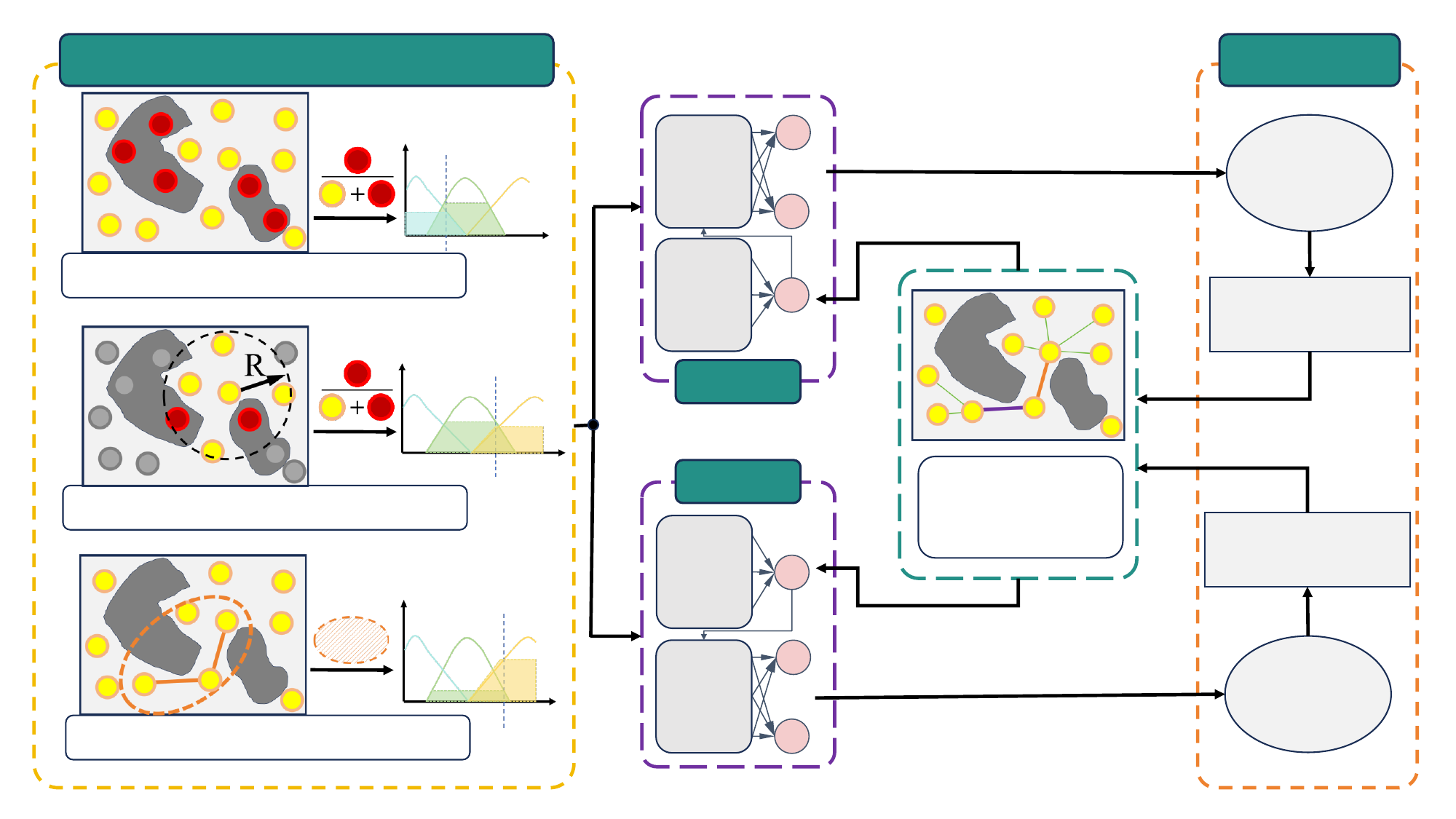}};
    \node[scale=0.95] at (-5.0,2.03) {\textcolor{white}{\footnotesize Environment Fuzzification}};
    \node at (0.9,2.03) {\textcolor{white}{\footnotesize Defuzzy}};

    \node at (-2.46,0.17) {\textcolor{white}{\footnotesize B-Net}};
    \node at (-2.45,-0.42) {\textcolor{white}{\footnotesize K-Net}};
    
    \node[scale=0.95] at (-5.22,0.78) {\footnotesize Global Invalid Ratio};
    \node[scale=0.95] at (-5.22,0.-0.59) {\footnotesize Local Invalid Ratio};
    \node[scale=0.95] at (-5.22,-1.93) {\footnotesize Lebesgue Measure};

    \node[scale=0.8] at (-2.63,1.4) {\footnotesize Actor};
    \node[scale=0.8] at (-2.63,0.65) {\footnotesize Critic};
    \node[scale=0.8] at (-2.63,-0.95) {\footnotesize Critic};
    \node[scale=0.8] at (-2.63,-1.68) {\footnotesize Actor};
    
    \node at (-2.1,1.45) {\footnotesize .};
    \node at (-2.1,1.40) {\footnotesize .};
    \node at (-2.1,1.35) {\footnotesize .};
    \node at (-2.1,-1.63) {\footnotesize .};
    \node at (-2.1,-1.68) {\footnotesize .};
    \node at (-2.1,-1.73) {\footnotesize .};

    \node[scale=0.8] at (-0.80,1.14) {\footnotesize Performance Valuation};
    \node[scale=0.8] at (-0.80,-1.32) {\footnotesize Performance Valuation};
    \node[scale=0.8] at (-0.80,-1.84) {\footnotesize Membership Degree K};
    \node[scale=0.8] at (-0.80,1.52) {\footnotesize Membership Degree B};

    \node[scale=0.65] at (-0.77,-0.44) {\footnotesize Learning-based};
    \node[scale=0.65] at (-0.77,-0.61) {\footnotesize Parameters};
    \node[scale=0.65] at (-0.77,-0.80) {\footnotesize Adoption};

    \node[scale=0.8] at (0.95,1.55) {\footnotesize B-};
    \node[scale=0.8] at (0.95,1.35) {\footnotesize Module};

    \node[scale=0.8] at (0.95,-1.55) {\footnotesize K-};
    \node[scale=0.8] at (0.95,-1.75) {\footnotesize Module};

    \node[scale=0.65] at (0.9,-0.85) {\footnotesize Defuzzification};
    \node[scale=0.65] at (0.9,0.55) {\footnotesize Defuzzification};

    \end{tikzpicture}
    \caption{Overview of the deep-fuzzy motion planning framework. The environmental information in configuration space is encoded by global, local invalid ratio and Lebesgue measure of the informed set. A learning-based fuzzy rule is applied, followed by a defuzzification to get crisp batchsize $\mathcal{B}$ and the number of neighbors $\mathcal{K}$.}
    \label{fig:framework}
    \vspace{-1.7em} 
\end{figure}
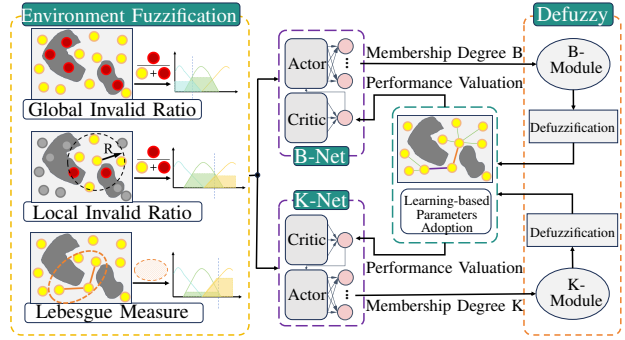
Recent work has focused on optimizing sampling strategies~\cite{zhang2025TASE} and nearest neighbor optimization~\cite{zhang2025apt}. Flexible Informed Trees (FIT*)~\cite{Zhang2024adaptive} dynamically adjust batch size to improve convergence, while Adaptive Prolated Trees (APT*)~\cite{zhang2025apt} employ prolated elliptical r-nearest neighbors to accelerate path search. However, these methods lack dynamic adaptation of batch size and neighbor selection~\cite{Zhang2024Elliptical}, potentially reducing accuracy and increasing search complexity. Moreover, reliance only on Coulomb's law limits their applicability.

Inspired by neural behavior, where the networks perceive a broader field of view in obstacle-sparse environments and navigate more directly, this work introduces Learning-Based Informed Trees (LIT*). LIT* integrates a deep fuzzy framework to address these challenges by leveraging invalid sampled states to encode environmental complexity and dynamically adjusting both sampling density and neighbor selection. By determining the number of sampled points and selected neighbors based on real-time conditions, LIT* enhances both planning efficiency and path quality.

\begin{figure}[t!]
    \centering
    \begin{tikzpicture}

    \node[inner sep=0pt] (russell) at (-2.5,0)
    {\includegraphics[width=0.49\textwidth]{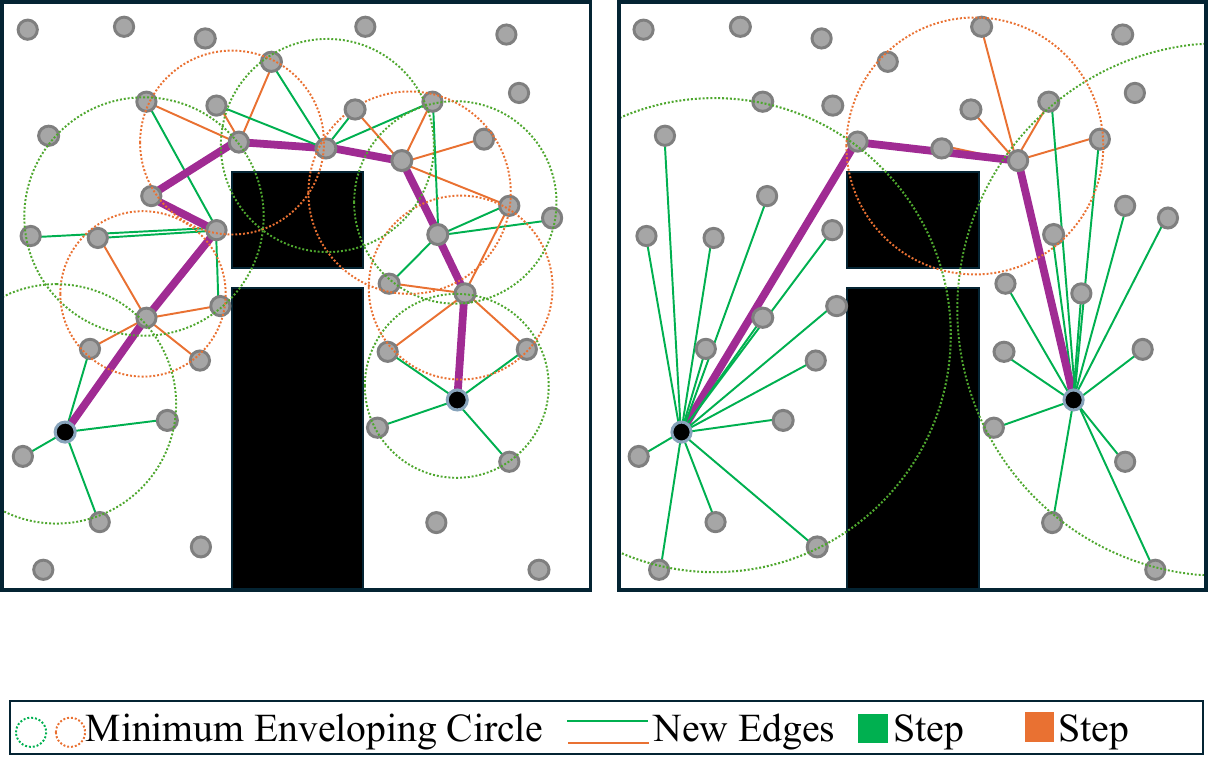}};
    \node at (0.15,-2.5) {\footnotesize  \textit{i}};
    \node at (1.45,-2.5) {\footnotesize  \textit{i+}1};

    \node at (-4.8,-1.9) {\footnotesize (a) Without learning-based $\mathcal{K}$};
    \node at (-0.3,-1.9) {\footnotesize (b) With learning-based $\mathcal{K}$};

    \node at (1.35,-0.3) { $\mathbf{x}_\textnormal{goal}$};
    \node at (-3.1,-0.3) { $\mathbf{x}_\textnormal{goal}$};
    \node at (-1.4,-0.6) { $\mathbf{x}_\textnormal{start}$};
    \node at (-6,-0.6) { $\mathbf{x}_\textnormal{start}$};


    \end{tikzpicture}
    \caption{Compare between with and without learning-based neighbor selection. These circles represent the minimum enclosing regions of selected neighbors at each step. Choosing $\mathcal{K}$ with a leaning-based method facilitates a more direct and cost-efficient path.}
    \label{fig:compare}
    \vspace{-1.7em} 
\end{figure}
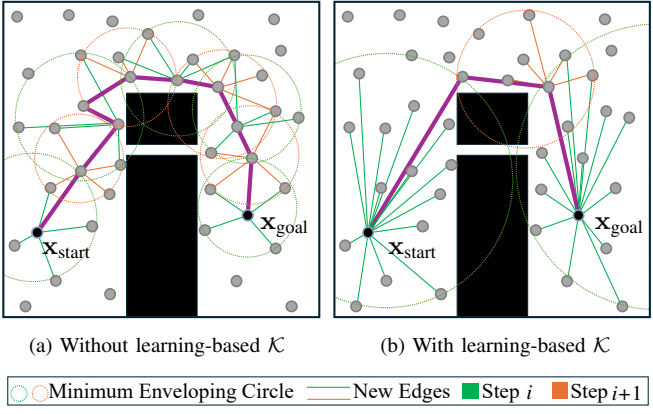

The main contributions of this work are:
\begin{itemize}
    \item \textit{A Fuzzy Reinforcement Learning Framework}: Encodes environmental complexity through \textit{global invalid rate}, \textit{local invalid rate}, and \textit{the Lebesgue measure} of the informed set, enabling adaptive motion planning.
    \item \textit{Learning-based batch size selection}: Dynamically adjusts batch size to balance exploration and exploitation.
    \item \textit{Learning-based neighbor selection}: Replaces the fixed $k$-nearest neighbor approach (Fig.~\ref{fig:compare}) with an adaptive strategy, improving sampling efficiency in both obstacle-dense and obstacle-sparse regions.
\end{itemize}

\section{Preliminaries and Problem formulation}
This section introduces the background of neighbor search and deep fuzzy systems as preliminaries. It then formulates the problem in the deep fuzzy framework, explaining how it improves motion planning efficiency and solution quality.

\subsection{Preliminaries}

\subsubsection{$R$-Nearest $\&$ K-Nearest Neighbors Search}
The $r$-nearest neighbors search defines connectivity based on a fixed radius around each state. A state $q$ is connected to all other states within a radius $r(q)$, The $k$-nearest neighbors search method connects each state $q$ to its $k$ closest states regarding Euclidean distance. Their can be formulated as:
\begin{equation}
\label{eqn: radius RNN}
    r(q) := \eta \left(2\left(1 + \frac{1}{n}\right){\left(\frac{\lambda(X_{\text{free}})}{\lambda\left(B_{1, n}\right)}\right) \left( \frac{\log(q)}{q}\right)}\right)^{\frac{1}{n}},
\end{equation}
where $\eta$ is a normalization constant, $n$ is the dimensionality of the space, $\lambda(\cdot)$ denotes the Lebesgue measure, $X_{\text{free}}$ is the free space, $B_{1,n}$ is the unit ball in $n$-dimensional space~\cite{strub2022adaptively}. 
\begin{equation}
\label{eq: knn}
    k(q) := \eta e \left( 1 + \frac{1}{n} \right) \log(q),
\end{equation}
where $\eta$ is a tuning parameter, $e$ is the base of the natural logarithm, and $n$ denotes dimension~\cite{andoni2009nearest}.


\subsubsection{Deep Fuzzy Systems}

Deep fuzzy systems combine fuzzy logic with learning methods.~\cite{FuzzyMachineLearning}. A general deep fuzzy framework consists of three key components: \textit{Fuzzification}: transforming raw input data into fuzzy membership values;~\textit{Fuzzy rules}: defines the relationships between fuzzy variables through a set of rules; and~\textit{Defuzzification}: applies fuzzy logic to derive crisp conclusions based on fuzzy rules.

\begin{algorithm}[t]
\caption{Learning-based Informed Trees (LIT*)}
\SetKwInOut{Input}{Input}
\SetKwInOut{Output}{Output}

\SetKwFunction{continueReverseSearch}{continueReverseSearch}
\SetKwFunction{continueForwardSearch}{continueForwardSearch}
\SetKwFunction{terminateCondition}{terminateCondition}
\SetKwFunction{prune}{prune}
\SetKwFunction{adaptiveBatchsize}{adaptiveBatchsize}
\SetKwFunction{improveApproximation}{improveApproximation}
\SetKwFunction{adaptiveNeighborSize}{adaptiveNeighborSize}
\SetKwFunction{sample}{sample}
\SetKwFunction{expand}{expand}
\SetKwFunction{sendToNN}{sendToNN}
\SetKwFunction{getMeasure}{getMeasure}
\SetKwFunction{deepFuzzy}{deepFuzzy}
\SetKwFunction{toIndex}{toIndex}
\SetKwFunction{localRatioCalc}{localRatioCalc}
\SetKwFunction{ratioCalc}{ratioCalc}

\DontPrintSemicolon
\small
\label{alg: g3t} 
\Input{$\text{Start vertex}~\mathbf{x}_\textnormal{start}$, \text{goal region}~$\textit{}{X}_\textnormal{goal}$}
\Output{$\text{Optimal feasible path}~\xi^*$}
\KwData{$\text{firstTime, solutionUpdate, expansionNeeded}$ \\ $ \text{trainingMode} \in \{\text{True}, \text{False}\}$, trained tensor$_\mathcal{B}$, tensor$_\mathcal{K}$}

$\textit{X}_{\textnormal{valid}} \gets \{\mathbf{x}_\textnormal{start}, \textit{}{X}_\textnormal{goal}\}$, $\mathbf{x}_{\textnormal{center,k}} \leftarrow \mathbf{x}_\textnormal{start}$, $\textit{X}_{\textnormal{invalid}} \gets \emptyset$, $E \mathcal \gets \emptyset$\\
$\textit{X}_{\textnormal{valid}}, \textit{X}_{\textnormal{invalid}} \xleftarrow{+} \sample(\mathcal{B}_{\textnormal{init}} ) $\\

\While{\textbf{not} $\terminateCondition()$}{

    $\rho_{\textnormal{global}} \leftarrow \ratioCalc(\textit{X}_{\textnormal{invalid}}, \textit{X}_{\textnormal{valid}})$ \Comment{Eq.~\ref{eq:ratioDefinition}} \\
    $X_{\textnormal{sample}} \leftarrow \textit{X}_{\textnormal{valid}} \cup \textit{X}_{\textnormal{invalid}}$ \\
    $\rho_{\textnormal{local,B}}, \rho_{\textnormal{local,K}} \leftarrow \localRatioCalc(X_{\textnormal{sample}}, \mathbf{x}_{\textnormal{center,k}}) $ \\
    $\lambda(X_{\hat{f}}) \leftarrow \getMeasure()$
    
    
    \If{\textnormal{solutionUpdated}}{
        \eIf{\textnormal{trainingMode}}{
            $\sendToNN(\rho_{\textnormal{global}}, \rho_{\textnormal{local,B}}, \lambda(X_{\hat{f}}))$\\
            \textcolor{purple}{$\mathcal{B} \leftarrow \deepFuzzy()$}
        }
        {
            $x, y, z \leftarrow \toIndex(\rho_{\textnormal{global}}, \rho_{\textnormal{local,B}}, \lambda(X_{\hat{f}}))$\\
            \textcolor{purple}{$\mathcal{B} \leftarrow \textnormal{tensor}_{\mathcal{B}}[x,y,z]$}
        }
    }
    get expand condition: expansionNeeded

    \If{\textnormal{expansionNeeded}}{
    \eIf{\textnormal{trainingMode}}{
            $\sendToNN(\rho_{\textnormal{global}}, \rho_{\textnormal{local,B}}, \lambda(X_{\hat{f}}))$\\
            \textcolor{purple}{$\psi_{\mathcal{K}} \leftarrow \deepFuzzy()$}
        }
        {
            $x, y, z \leftarrow \toIndex(\rho_{\textnormal{global}}, \rho_{\textnormal{local,K}}, \lambda(X_{\hat{f}}))$\\
            \textcolor{purple}{$\psi_{\mathcal{K}} \leftarrow \textnormal{tensor}_{\mathcal{K}}[x,y,z]$}
        }
    $\mathcal{K} \leftarrow \eta \cdot \psi_{\mathcal{K}} \cdot (1 + \frac{1}{n_{\textnormal{dimension}}}) \log(|X_{\textnormal{valid}}|)$ \Comment{Eq.~\ref{eq:newknn}}\\
    $E \xleftarrow{+} \expand(x_{\textnormal{p}}, \mathcal{K})$
    }

    $\prune(\textit{X}_{\textnormal{sample}}, \sample(\mathcal{B}))$
    }
\Return $\xi^{*}$
\end{algorithm}
\begin{algorithm}[h]
\caption{LIT* - Local ratio calculation}
\KwIn{$X_{\textnormal{sample}}$, $\mathbf{x}_{\textnormal{center,k}}$}
\KwOut{$\rho_{\textnormal{local,B}}, \rho_{\textnormal{local,K}}$}
\SetKwFunction{sendMapInformationToNN}{sendMapInformationToNN}
\SetKwFunction{getValidInvalid}{getValidInvalid}
\SetKwFunction{getStatesOnPath}{getStatesOnPath}
\SetKwFunction{ratioCalc}{ratioCalc}
\SetKwFunction{caucLocalRadius}{caucLocalRadius}

$X_{\smash{\textnormal{valid,r}}} \mathcal \gets \emptyset, X_{\smash{\textnormal{invalid,r}}} \mathcal \gets \emptyset, X_{\smash{\textnormal{center,b}}} \mathcal \gets \emptyset$ \\
$X_{\smash{\textnormal{center,b}}} \leftarrow \getStatesOnPath(\xi_{\textnormal{current}})$ \\
$r \leftarrow \caucLocalRadius()$ \Comment{Eq.~\ref{eqn: radius RNN}}\\
\ForEach{ $\mathbf{x} \in X_{\smash{\textnormal{center,b}}}$}{
    $X_{\smash{\textnormal{valid,r}}}, X_{\smash{\textnormal{invalid,r}}} \leftarrow \getValidInvalid(\mathbf{x}, r)$\\
    $\rho_{\textnormal{local,B}} \xleftarrow{+} \ratioCalc(\textit{X}_{\textnormal{invalid,r}}, \textit{X}_{\textnormal{valid,r}})$ \Comment{Eq.~\ref{eq:ratioDefinition}}\\ 
}
$\rho_{\textnormal{local,B}} \leftarrow \rho_{\textnormal{local,B}} / |X_{\smash{\textnormal{center,b}}}|$ \\
$X_{\smash{\textnormal{valid,r}}}, X_{\smash{\textnormal{invalid,r}}} \leftarrow \getValidInvalid(\mathbf{x}_{\smash{\textnormal{center,k}}}, r)$\\
$\rho_{\textnormal{local,K}} \leftarrow \ratioCalc(\textit{X}_{\textnormal{invalid,r}}, \textit{X}_{\textnormal{valid,r}})$ \\
\Return $\rho_{\textnormal{local,B}}, \rho_{\textnormal{local,K}}$
\end{algorithm}

This work replaces the fuzzy rule component with a reinforcement learning algorithm, \textit{Deterministic Policy Gradient (DDPG)} architecture, as illustrated in Fig.~\ref{fig:framework}. It generates a deterministic policy as the fuzzy rule and facilitates seamless integration into subsequent planning stages.

\subsection{Problem Formulations}
\subsubsection{Problem Formulation 1 (Optimal Planning)}
We define the motion planning problem within the framework of sampling-based methods~\cite{karaman2011sampling}. The state space is denoted as \( X \subseteq \mathbb{R}^n \), where obstacle regions occupy \( X_{\text{invalid}} \), the set of valid, collision-free states is defined as \(X_{\text{valid}}\), and the goal states is \(X_{\text{goal}} \subset X_{\text{valid}}\).
The motion planning problem is defined to compute a path \( \xi : [0,1] \to X_{\text{valid}} \) such that \( \xi(0) = \mathbf{x}_{\text{init}} \) and \( \xi(1) \in X_{\text{goal}}\). The optimal path denoted as \(\xi^{*}\); \(\Xi\) denotes the set of all nontrivial paths. The path planning problem can be described as:
\begin{align*}
    & \xi^* = \arg \min_{\xi \in \Xi} c(\xi) \\
    \text{s.t.} \quad 
    & \xi(0) = \mathbf{x}_{\text{init}}, \xi(T) \in X_{\text{goal}}, \xi(t) \in X_{\text{valid}}(t), \forall t \in [0,1],
\end{align*}
where \(c(\xi): \Xi \mapsto \mathbb{R}_{\geq 0}\) is the cost function of each feasible path, and the optimal cost is $c^*$.

\subsubsection{Problem Formulation 2 (Fuzzy-DDPG)}  
Beyond solving the motion planning problem, this work aims to improve the solution quality using the deep fuzzy framework, where reward functions \(\mathcal{R}\) are designed to guide learning. During training, the objective is to maximize reward, ensuring efficient learning of key parameters.

The reinforcement learning observation \( \textbf{s}_t \) is fuzzified global, local ratio and Lebesgue measure. 
Each state \( \textbf{s}_t \) is associated with an action \( \textbf{a}_t \in \mathbb{R}^3 \). After defuzzification , a crisp number of batchsize \(\mathcal{B} \in \mathbb{N}^{+} \cap [20, 200]\) and a factor of neighbor selection \( \psi_{\mathcal{K}} \in \mathbb{R}^+ \cap [3.0, 15.0]  \) will be applied in the following algorithm. The limits of batchsize are based on SOTA algorithms' upper and lower limits, while the factor's limits consider the connectivity.


The $\textit{Fuzzy-DDPG}$'s task is to determine an optimal policy \(\bm{\pi}: \mathbb{S} \to \mathbb{A}\), where \( \mathbb{S} \) is the state space, and \( \mathbb{A} \) is the action space. The learned policy \( \bm{\pi} \) aims to accelerate convergence and reduce solution cost, which are valued by the reward function $\mathcal{R}$. Formally, the problem can be described as:
\begin{align*}
    & \argmax_{\bm{\pi}} \mathcal{R}  \\
    \text{s.t.} \quad 
    & \tilde{\mathcal{B}} = \bm{\pi}_{B}(\mathbf{s}_t ; \bm{W}_{B}), \tilde{\psi_{\mathcal{K}}} = \bm{\pi}_{K}(\mathbf{s}_t; \bm{W}_{K}), 
\end{align*}
where \( \bm{W}_{B} \) and \( \bm{W}_{K} \) represent the parameters of the neural network trained via reinforcement learning.

\begin{figure}[t]
    \centering
   
    \begin{tikzpicture}
        \node[inner sep=0pt] (russell) at (0.0,0.0)
        {\includegraphics[width=0.49\textwidth]{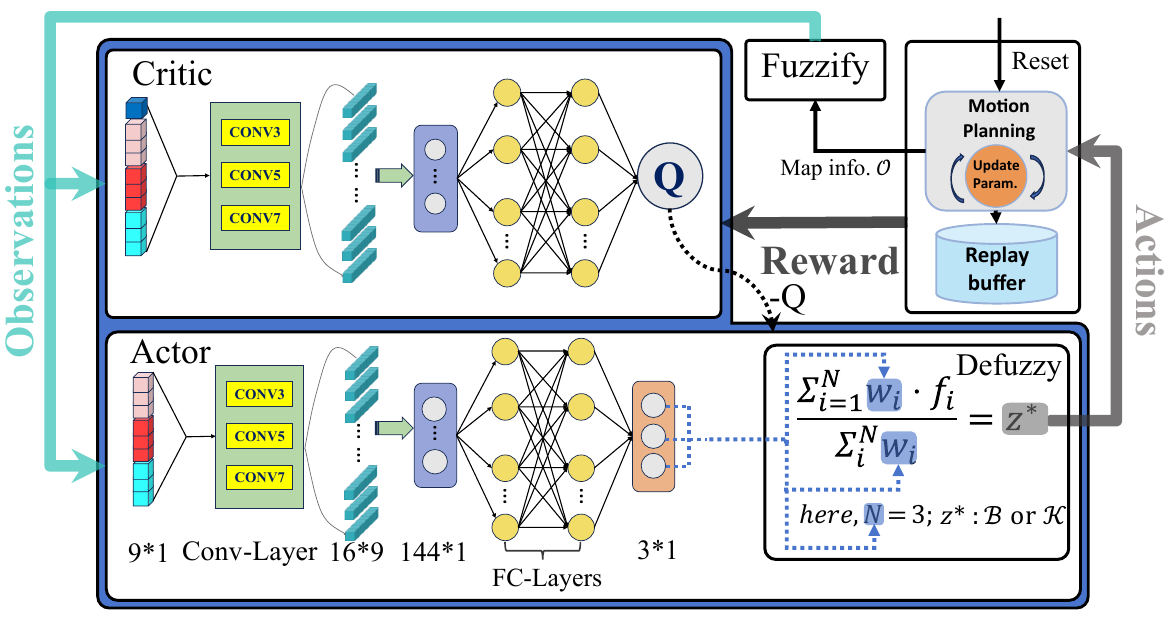}};
    \end{tikzpicture}
    \caption{Illustration of the $\textit{Fuzzy-DDPG}$ architecture. The Actor and Critic net have the same structure of convolutional layers, flattened layers, and fully connected layers. The essential difference is the output generated by defuzzification. }
    \label{fig:actorcritic}
    \vspace{-1.3em} 
\end{figure}


\section{Learning-Based Informed Trees (LIT*)}

\textit{Notation:}
In the Fuzzy-DDPG framework, observation $\textbf{s}_t$ encodes the map information through a fuzzification process; action \(\textbf{a}_t\) represents the membership values in the defuzzification phase;  $\mathcal{E}$  denotes the set of factors involved in reward calculation. Each step of reward $r_t$ is calculated based on $\mathcal{E}$. The policy network (Actor) and the value network (Critic) are denoted as $\pi_{\theta}$ and $Q_{\phi}$, their loss is $\mathcal{L}_{\theta}$ and $\mathcal{L}_{\phi}$.  To encode the map information, this work else takes invalid states $X_\textnormal{invalid}$ into account and defines the invalid ratio as:
\begin{equation}
    \rho = \frac{|X_\textnormal{invalid}|}{|X_\textnormal{valid}| + |X_\textnormal{invalid}|},
    \label{eq:ratioDefinition}
\end{equation}
where the $|\cdot|$ denotes the size of corresponded set.
\begin{algorithm}[t]
\caption{LIT* - Fuzzy-DDPG}
\label{alg:fuzzy-ddpg}

\SetKwFunction{startLIT}{startLIT$^{*}$}
\SetKwFunction{getMapInfo}{getMapInfo}
\SetKwFunction{reset}{getMapInfo}
\SetKwFunction{resetEnvironment}{resetEnvironment}
\SetKwFunction{fuzzify}{fuzzify}
\SetKwFunction{defuzzify}{defuzzify}
\SetKwFunction{getPlannerInfo}{getPlannerInfo}
\SetKwFunction{size}{size}
\SetKwFunction{sample}{sample}
\SetKwFunction{rewardFunction}{rewardFunction}
\SetKwFunction{lossOfCritic}{lossOfCritic}
\SetKwFunction{updateCritic}{updateCritic}
\SetKwFunction{updateActor}{updateActor}
\SetKwFunction{size}{size}

\KwIn{map information $\mathcal{O}$ }
\KwOut{trained policy $\pi_{\theta}$}
\KwData{reward info. $\mathcal{E}$, replay buffer $\mathcal{D}$, minibatch $\bm{b}$}

Initialize  $\pi_{\theta}$ ,  $Q_{\phi}$ , and  $\mathcal{D}$ \\

\For{$episode \gets 1$ \KwTo $N$}{
    $\mathcal{O}_{0} \leftarrow \resetEnvironment()$ \\
    \textcolor{purple}{$\textbf{s}_{0} \leftarrow \fuzzify(\mathcal{O}_{0})$} \\

    \While{not done}{
        \textcolor{purple}{$a_t \leftarrow \defuzzify(\pi_{\theta}(s_t))$} \\
        $(\mathcal{O}_{t+1}, \mathcal{E}_t, done) \leftarrow \getPlannerInfo(a_t)$ \\
        \textcolor{purple}{$\textbf{s}_{t+1} \leftarrow \fuzzify(\mathcal{O}_{t+1})$} \\
        $r_t \leftarrow \rewardFunction(\mathcal{E}_t)$
        $(\textbf{s}_t, a_t, r_t, s_{t+1}) \rightarrow \mathcal{D}$ \\

        \If{$\mathcal{D}.\size() > m$}{
            $\bm{b} \gets \sample(m)$ \\
            $\mathcal{L}_\phi \gets \frac{1}{m} \sum\limits_{\bm{b}} \left( r + \gamma Q_{\phi'}(s', a') - Q_{\phi}(s, a) \right)^2$ \\
            $\mathcal{L}_\theta \gets - \frac{1}{m} \sum\limits_{\bm{b}} Q_{\phi}(s, \pi_\theta(s))$ \\
            apply soft-update using $\mathcal{L}_\phi$ and $\mathcal{L}_\theta$ \\
        }

    }
}
\Return $\pi_{\theta}$

\end{algorithm}

\subsection{Architecture of \textit{Fuzzy-DDPG}}
LIT* deconstructs the solving process from a microscopic algorithmic perspective, simulating a network's field of view and autonomously selects optimal actions based on the current environment. Fig.~\ref{fig:actorcritic} illustrates the architecture of \textit{Fuzzy-DDPG}. The process begins with fuzzifying three inputs, \textit{Fuzzy-DDPG} then iteratively trains the Actor and Critic. Finally, leveraging the trained Actor, the system computes membership degrees and applies defuzzification to generate a continuous action based on the learned policy.
Specificlly, for B-Net the defuzzied $z^*$ is batchsize $\mathcal{B}$ and for K-net the $z^*$ is $\psi_{\mathcal{K}}$, and $\mathcal{K}$ is calculated as:
\begin{equation}
\label{eq:newknn}
    k(q) := \eta e \cdot \psi_\mathcal{K} \cdot \left( 1 + \frac{1}{n} \right) \log(q),
\end{equation}
\begin{figure*}[t!]
    \centering
    \begin{tikzpicture}
    \footnotesize
    \node[inner sep=0pt] (russell) at (-8.0,0)
    {\includegraphics[width=0.24\textwidth]{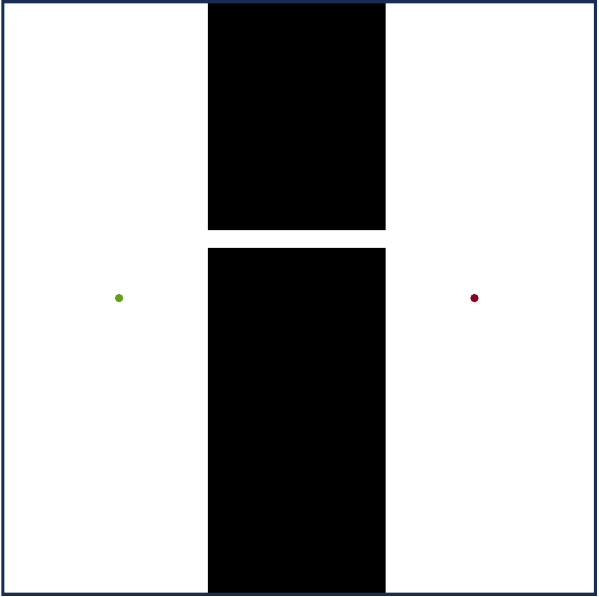}};
    \node[inner sep=0pt] (russell) at (-3.5,0)
    {\includegraphics[width=0.24\textwidth]{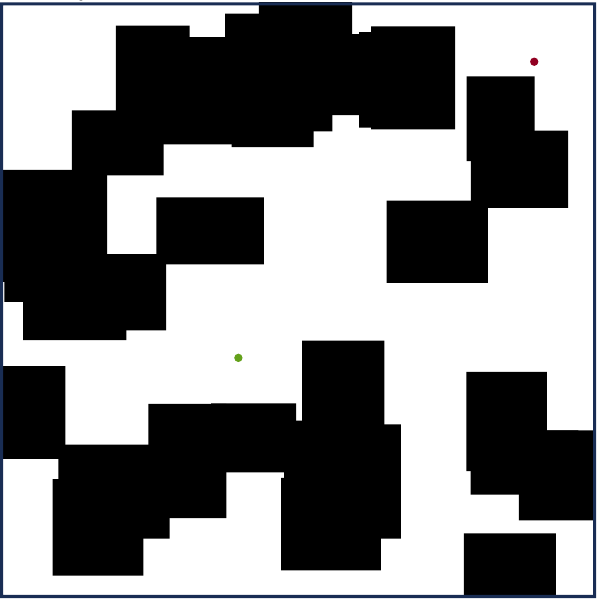}};
    \node[inner sep=0pt] (russell) at (1.0,0)
    {\includegraphics[width=0.242\textwidth]{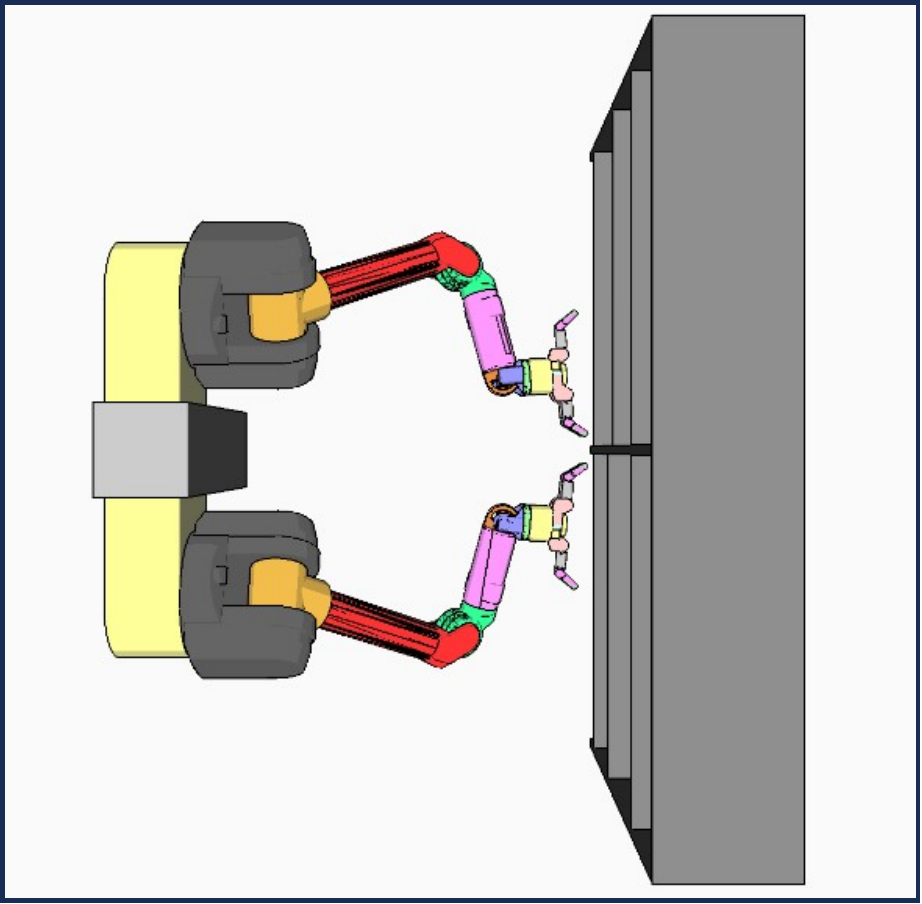}};
    \node[inner sep=0pt] (russell) at (5.5,0)
    {\includegraphics[width=0.242\textwidth]{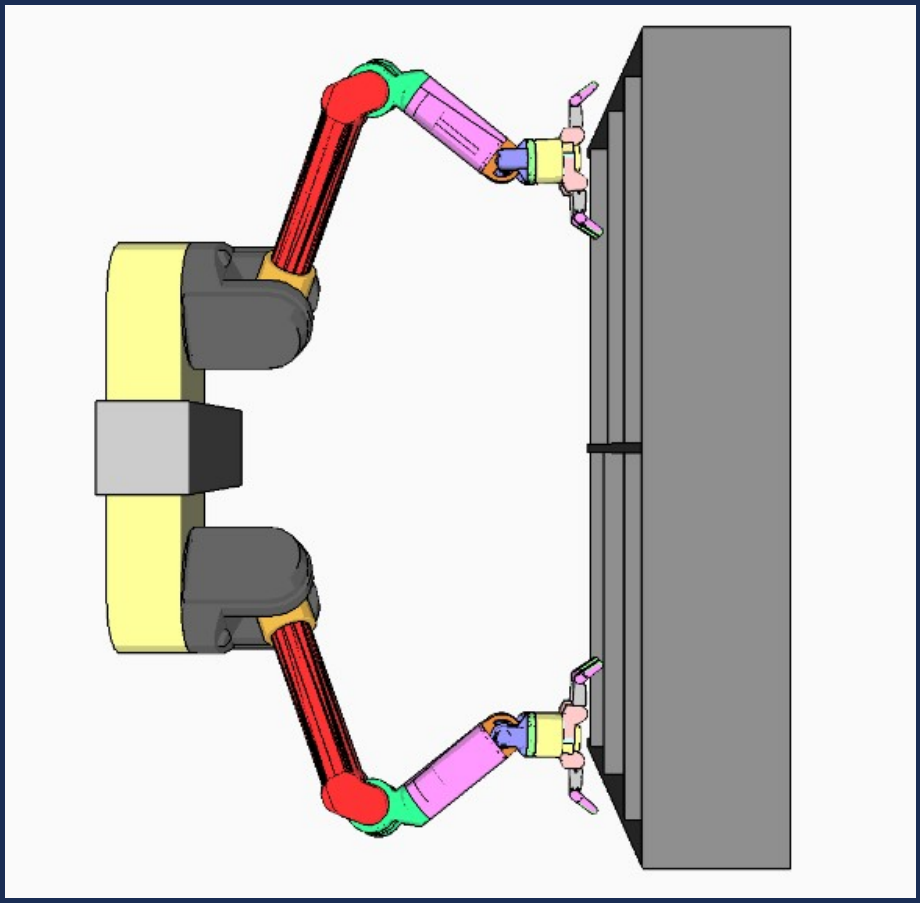}};
    
    \node at (-8.0,-2.4) {\small (a) Narrow Passage (NP)};
    \node at (-3.5,-2.4) {\small (b) Random Rectangles (RR)};
    \node at (1.0,-2.4) {\small (c) Start configuration};
    \node at (5.5,-2.4) {\small (d) Goal configuration};

    \node at (-9.3,0.3) { $\mathbf{x}_\textnormal{start}$};
    \node at (-6.7,0.3) { $\mathbf{x}_\textnormal{goal}$};

    \node at (-4.0,-0.25) { $\mathbf{x}_\textnormal{start}$};
    \node at (-1.8,1.9) { $\mathbf{x}_\textnormal{goal}$};
    
    \end{tikzpicture}
    \caption{Illustrates the narrow passage (a) and random rectangles (b) tests in Planner Developer Tools (PDT)~\cite{gammell2022planner}. Fig. (c) and (d) illustrate the dual-Barrett Whole-Arm Manipulator-$\mathbb{R}^{14}$ in the Open Robotics Automation Virtual Environment (OpenRAVE)~\cite{Diankov2008OpenRAVEAP}. (c) shows the start configuration as it picks up objects from the bottom, (d) illustrates the goal configuration where the object is placed on the top shelf.}
    \label{fig: simulation}
    \vspace{-1.0em}
\end{figure*}
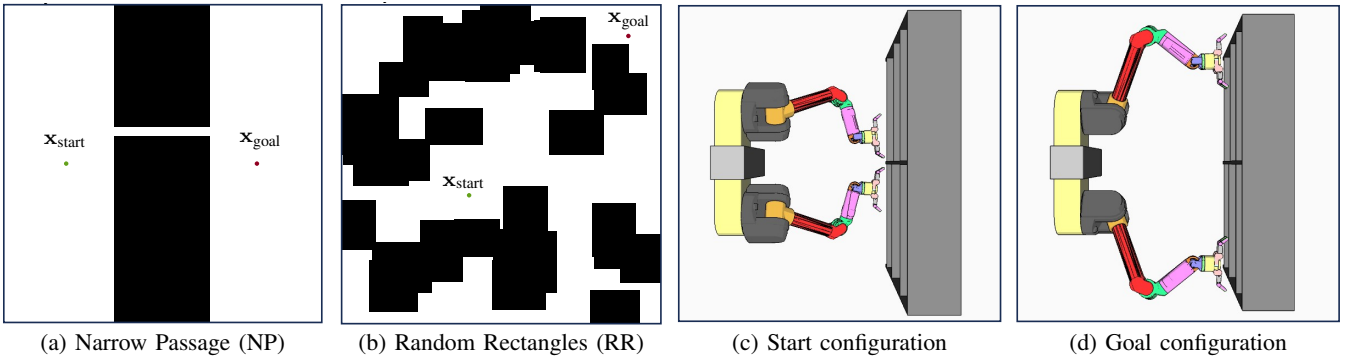
\begin{figure}[t!]
    \centering
    \begin{subfigure}[b]{0.24\textwidth}%
        \centering
        \begin{tikzpicture}
            \node[anchor=south west] (img) at (0,0) {\includegraphics[width=\textwidth]{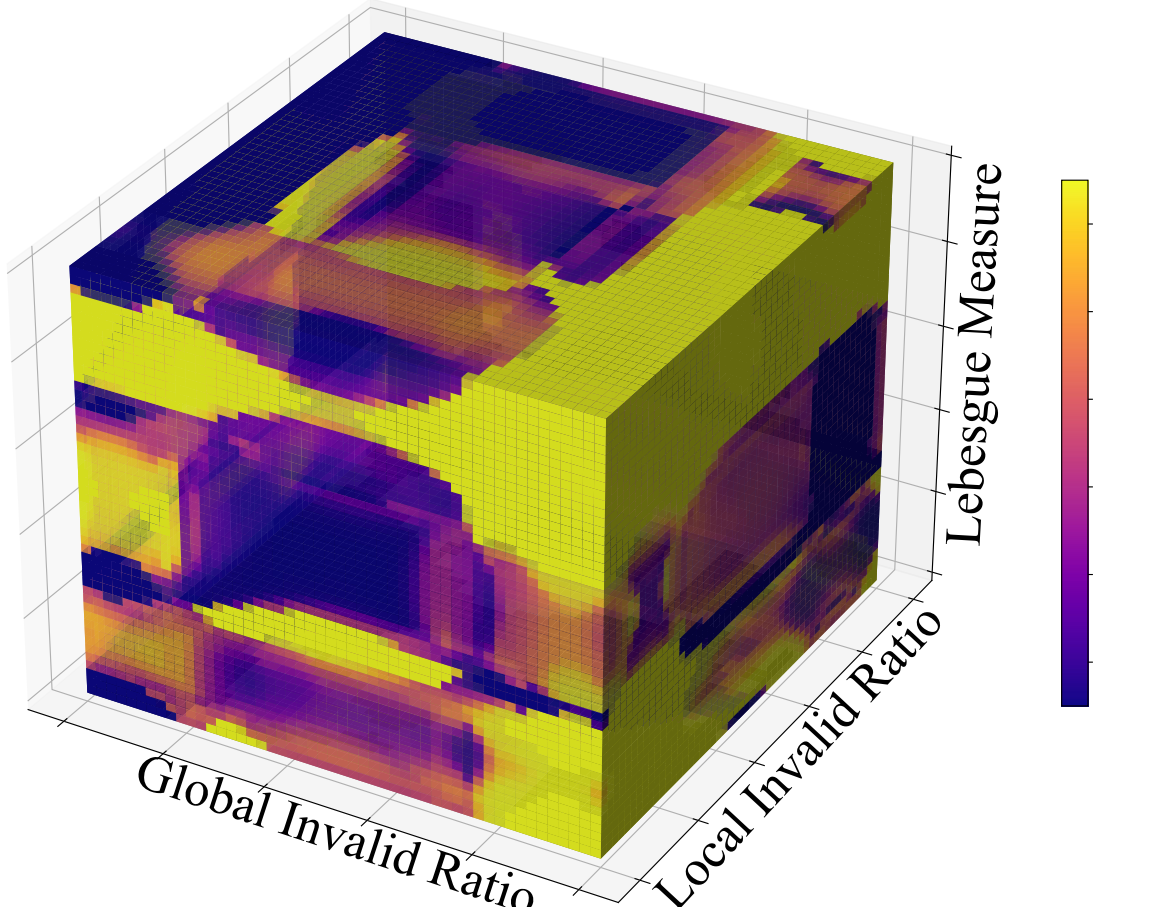}};

            
            \begin{scope}[x={(img.south east)},y={(img.north west)}] 
                \node[scale=0.7] at (0.08, 0.18) {\footnotesize 0.0};  
                \node[scale=0.7] at (0.52, 0.02) {\footnotesize 1.0};  
                \node[scale=0.7] at (0.83, 0.36) {\footnotesize 0.0}; 
                \node[scale=0.7] at (0.81, 0.85) {\footnotesize 1.0}; 
                \node[scale=0.7] at (0.92, 0.22) {\footnotesize 3.0};  
                \node[scale=0.7] at (0.92, 0.82) {\footnotesize 15.0}; 
            \end{scope}
        \end{tikzpicture}
        \caption{Trained Tensor of $\mathcal{K}$}
        \label{fig:compare1}
    \end{subfigure}%
    \hfill%
    \begin{subfigure}[b]{0.24\textwidth}%
        \centering
        \begin{tikzpicture}
            \node[anchor=south west] (img) at (0,0) {\includegraphics[width=\textwidth]{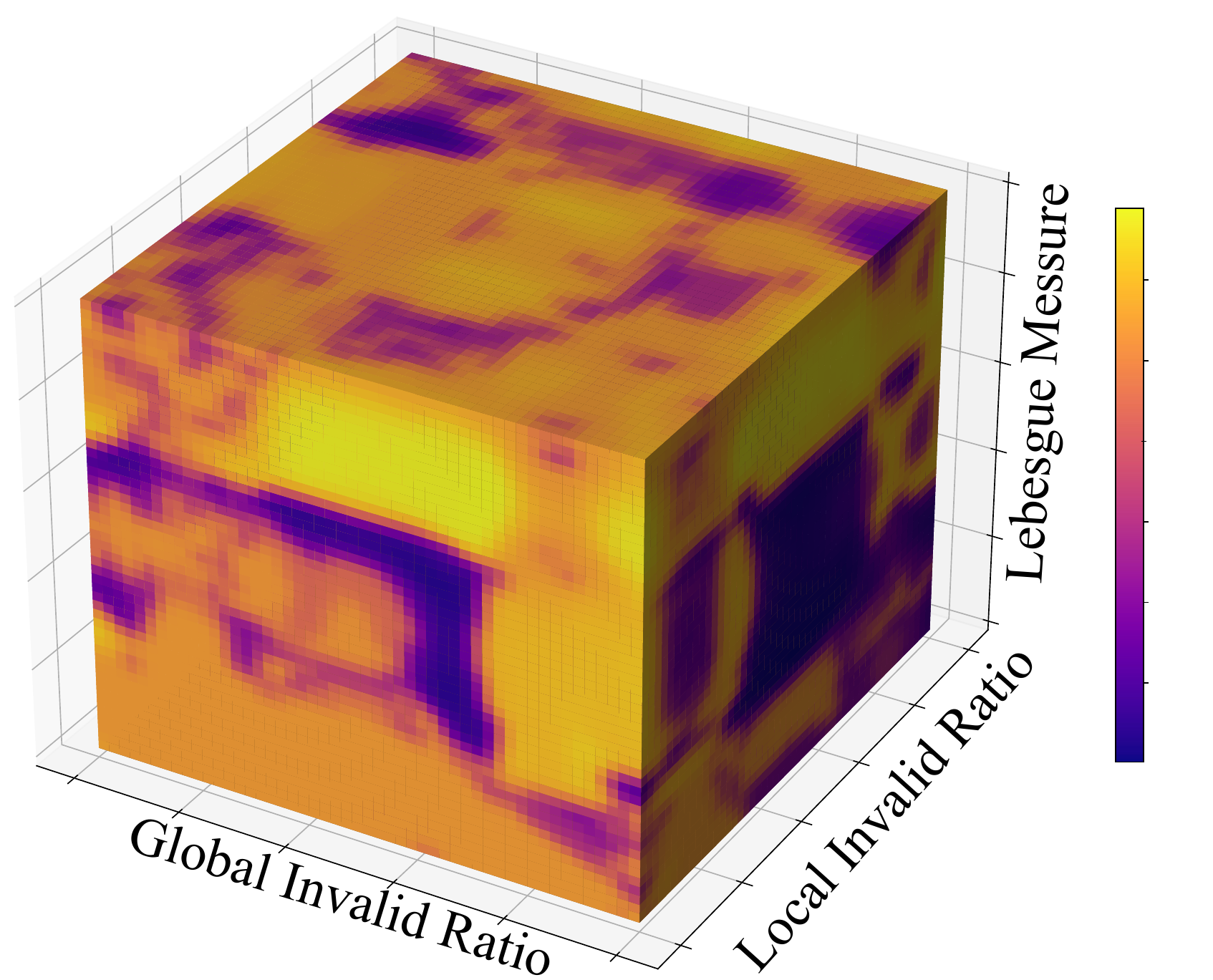}};

            
            \begin{scope}[x={(img.south east)},y={(img.north west)}] 
                \node[scale=0.7] at (0.08, 0.18) {\footnotesize 0.0};  
                \node[scale=0.7] at (0.52, 0.02) {\footnotesize 1.0};  
                \node[scale=0.7] at (0.83, 0.39) {\footnotesize 0.0}; 
                \node[scale=0.7] at (0.81, 0.85) {\footnotesize 1.0}; 
                \node[scale=0.7] at (0.92, 0.19) {\footnotesize 71};  
                \node[scale=0.7] at (0.92, 0.80) {\footnotesize 139}; 
            \end{scope}
        \end{tikzpicture}
        \caption{Trained Tensor of $\mathcal{B}$}
        \label{fig:compare2}
    \end{subfigure}%
    \caption{Visulization of the $\mathcal{K}$-tensor and $\mathcal{B}$-tensor The three axes of the tensor represent the global invalid ratio, the local invalid ratio, and the Lebesgue measure of the informed set. Each pair of these coordinates uniquely determines a specific value of $\mathcal{K}$ or $\mathcal{B}$.}
    \label{fig:tensor}
    \vspace{-1.0em}
\end{figure}
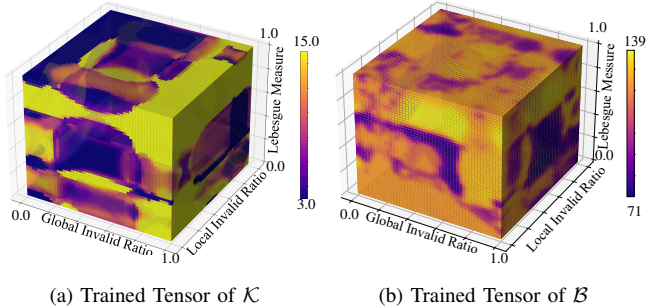
\subsubsection{Inputs Fuzzification}
This work extracts three key parameters to encode the map: $\rho_{\textnormal{global}}$, $\rho_{\textnormal{local}}$, and $\lambda(X_{\hat{f}})$. To further enhance the expressiveness of the input representation, this work fuzzifies these inputs into the \textit{DDPG} neural network to improve its feature representation capability.


This work selects three fuzzy sets and Gaussian membership functions to balance expressiveness and computational complexity. For each input in $\mathcal{O} = \{ \rho_{\text{global}}, \rho_{\text{local}}, \lambda(X_{\hat{f}}) \}$, this study employs fuzzy sets for modeling and designs three fuzzy sets to describe their characteristics:  \textit{S} (sparse), \textit{M} (medium), and \textit{D} (dense).

\begin{equation}
    \mu_{\mathcal{O},i}(\rho_{\text{global}}) = \exp\left(-\frac{(\rho_{\text{global}} - \vartheta_{\text{global},i})^2}{2\sigma_{\text{global},i}^2}\right),
\end{equation}

Correspondingly, each element in $\mathcal{O}$ has three membership functions (\( i, j, k \in \{1,2,3\} \)), corresponding to different levels of its fuzzy feature. The parameters \( \vartheta_{|\cdot|} \) and \( \sigma_{|\cdot|} \) control the shape and position of each membership function. Then, the final fuzzy mapping can be represented as a \( 9 \times 1 \) vector:
\begin{equation}
\bm{i} = [\mu_{\mathcal{O}_{1},1}(x_{1}), \mu_{\mathcal{O}_{1},2}(x_{1}), ... , \mu_{\mathcal{O}_{3},3}(x_{3})]^T,
\label{input}
\end{equation}
where \( x_1, x_2, x_3 \) correspond to the three components of the input \( \mathcal{O}\). This fuzzy logic mapping transforms the original input data into a continuously fuzzy representation, making it more effectively processable by the neural network.

\subsubsection{Actor and Critic}
In \textit{Fuzzy-DDPG}, there are two primary networks: the Actor 
\(A_{\theta}(\mathbf{s})\) and the Critic \(Q_{\phi}(\mathbf{s, a})\). The Actor network takes a 9-dimensional input and passes it through a convolutional layer with kernel sizes 3, 5 and 7, and outputs 9 channels, followed by a fully connected multi-layer perceptron (MLP). The MLP consists of five hidden layers with neuron sizes (64, 128, 128, 64, 32) activated by ReLU. The final output layer has 3 neurons, corresponding to the number of defuzzification-membership functions, and applies a TSK defuzzification~\cite{softmaxNeuralNetwork} for discrete and continuous actions. The Critic network follows a similar architecture but takes the concatenation of state \(\textbf{s}\) and defuzed-result as input. The output layer is a single neuron representing the estimated Q-value of current (\(\textbf{s}\), \(\textbf{a}\)) pair.

\subsubsection{Defuzzification}

This work select the Takagi-Sugeno-Kang (TSK) defuzzification~\cite{TSKDefuzzy} method due to its effectiveness in modeling complex nonlinear systems and its widespread use in control and decision-making applications. It employs a weighted sum approach, allowing a smooth transition between rules and enabling more adaptive decision-making. Specifically, the defuzzified output is computed as:

\begin{equation}
    z^* = \frac{\sum_{i=1}^{N} w_i f_i}{\sum_{i=1}^{N} w_i},
\end{equation}

where \( w_i \) represents the weight of the \( i \)-th fuzzy rule, and \( f_i \) is a constant functions to control the output range.







\subsection{Training of Fuzzy-DDPG}

Deep Deterministic Policy Gradient (DDPG) is an actor-critic reinforcement learning algorithm designed for continuous control problems. The key feature of DDPG is its ability to optimize policies over a continuous action space by leveraging an off-policy training framework with target networks and soft updates.

The Bellman equation for the critic network is defined as:
\begin{equation}
    Q_{\phi}(\mathbf{s}_t, \mathbf{a}_t) = r_t + \gamma Q_{\phi'}(\mathbf{s}_{t+1}, A_{\theta'}(\mathbf{s}_{t+1})),
\end{equation}
where: \( r_t \) is the reward at time step \( t \),  \( Q_{\phi'} \) and \( A_{\theta'} \) are the target networks for the critic and actor, respectively, \( \gamma \in [0,1] \) is the discount factor.

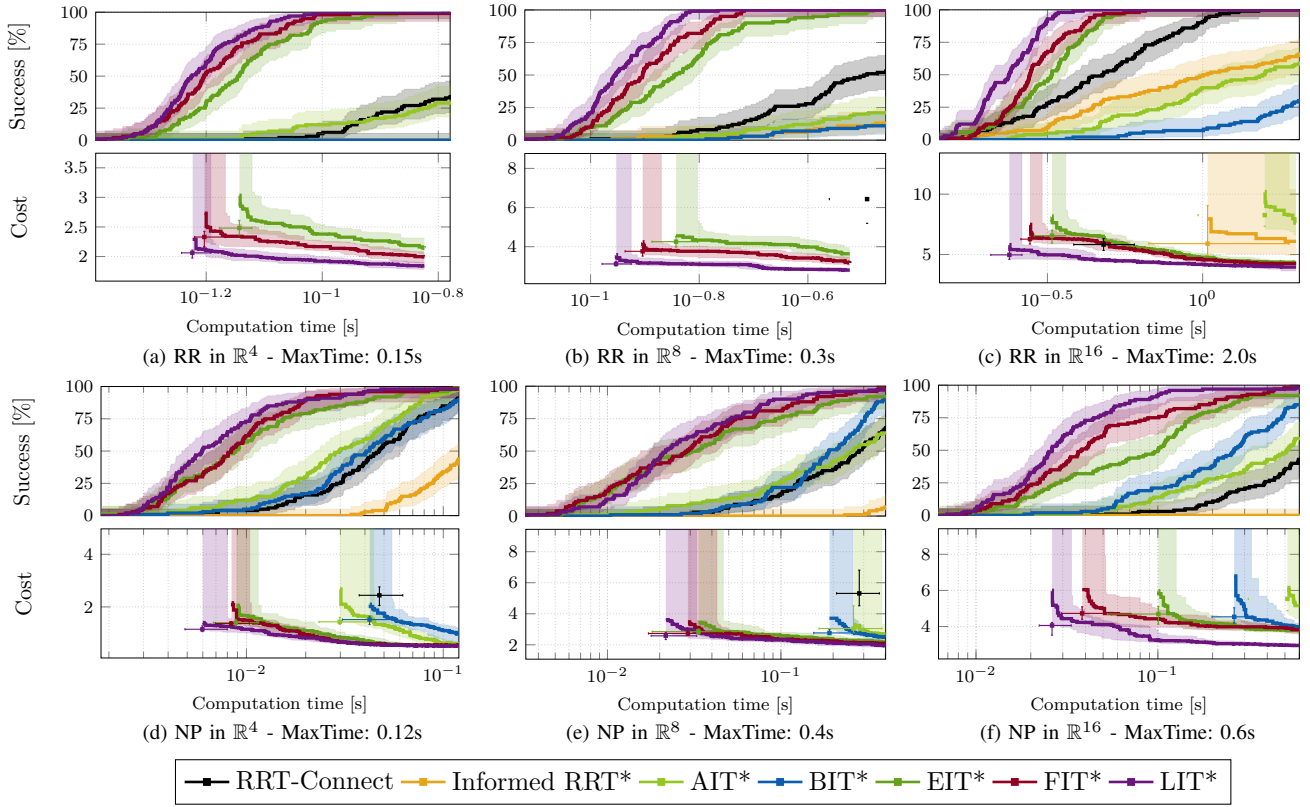
\begin{figure*}[t!]
    \centering
    \begin{tikzpicture}
    \node[inner sep=0pt] (russell) at (-6.2,8)
    {\includegraphics[width=0.35\textwidth]{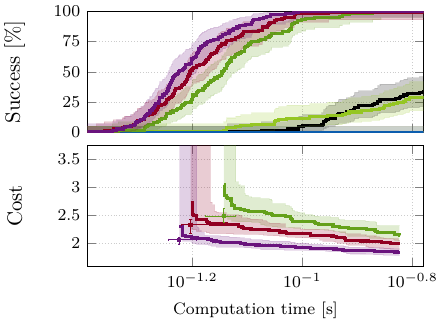}};
    \node[inner sep=0pt] (russell) at (-0.2,8)
    {\includegraphics[width=0.30\textwidth]{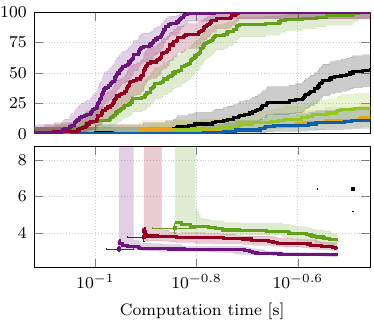}};
    \node[inner sep=0pt] (russell) at (5.3,8)
    {\includegraphics[width=0.30\textwidth]{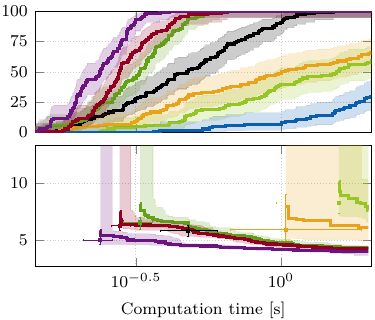}};

    \node[inner sep=0pt] (russell) at (-6.2,3)
    {\includegraphics[width=0.34\textwidth]{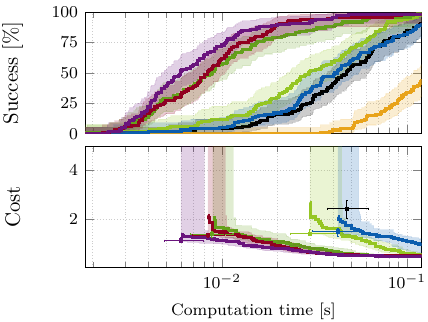}};
    \node[inner sep=0pt] (russell) at (-0.2,3)
    {\includegraphics[width=0.30\textwidth]{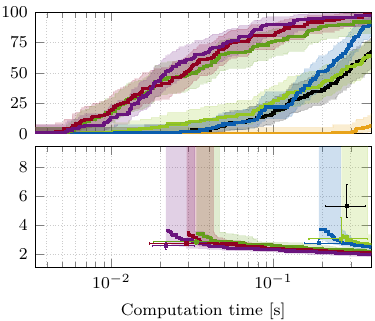}};  
    \node[inner sep=0pt] (russell) at (5.3,3){\includegraphics[width=0.30\textwidth]{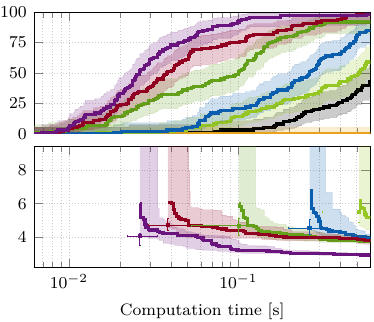}};

    \node[inner sep=0.0pt] (russell) at (0.0,-0.12){\includegraphics[width=0.8\textwidth]{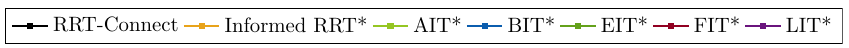}};
    
    \node at (-5.6,5.5) {\footnotesize (a) RR in $\mathbb{R}^4$ - MaxTime: 0.15s};
    \node at (-0.05,5.5) {\footnotesize(b) RR in $\mathbb{R}^8$ - MaxTime: 0.3s};
    \node at (5.5,5.5) {\footnotesize(c) RR in $\mathbb{R}^{16}$ - MaxTime: 2.0s};
    \node at (-5.6,0.5) {\footnotesize (d) NP in $\mathbb{R}^4$ - MaxTime: 0.12s};
    \node at (-0.05,0.5) {\footnotesize(e) NP in $\mathbb{R}^8$ - MaxTime: 0.4s};
    \node at (5.5,0.5) {\footnotesize(f) NP  in $\mathbb{R}^{16}$ - MaxTime: 0.6s};

    \end{tikzpicture}
    \vspace{-0.1em} 
    \caption{Detailed experimental results from Section~\ref{sec:experi} are presented above. Fig. (a), (b), and (c) depict test benchmark random rectangle outcomes in $\mathbb{R}^4$, $\mathbb{R}^8$ and $\mathbb{R}^{16}$, respectively. Panel (d) showcases ten narrow passage experiments in $\mathbb{R}^4$, while panels (e) and (f) demonstrate in $\mathbb{R}^8$ and $\mathbb{R}^{16}$. In the cost plots, boxes represent solution cost and time, with lines showing cost progression for planners (unsuccessful runs have infinite costs). Error bars provide nonparametric 99\% confidence intervals for solution cost and time.}
    \label{fig: result}
    \vspace{-1.8em}
\end{figure*}

\subsubsection{Reward Definition}
This work defines different reward functions for two sub-networks: BatchSize-Net (B-Net) and NearestK-Net (K-Net) as in Fig.~\ref{fig:framework}. Processing more information each step in an obstacle-free space can lead to more direct solutions and increase computational effort. Therefore, the reward function must account for key parameters to balance efficiency and solution quality.

\paragraph{BatchSize-Net Reward}
The reward function for B-Net is defined as:
\begin{equation}
\mathcal{R}_{B} = \alpha_B \cdot \frac{\kappa}{t} + \beta_B \cdot \frac{\kappa}{c(\xi)} -\gamma_B \cdot n_{\text{update}},
\label{rewardB}
\end{equation}
where: \( \alpha_B \), \( \beta_B \), and \( \gamma_B \) are scaling factors to ensure that the corresponding terms remain within a reasonable range, preventing them from becoming excessively large or small; t denoted the time cost for each solution update; \( n_{\text{update}} \) represents the number of solution updates, \( \gamma_B \) is a scale parameter for \( n_{\text{update}} \), \( \kappa\) is a decay function about \( n_{\text{update}} \), which is given by:
  \[
  \kappa(n_{\text{update}}) = \max \left( \nu_{\min}, \nu \cdot \log_2 (6.8 - n_{\text{update}}) \right).
  \]

This reward function encourages rapid convergence by assigning high rewards in early updates if a solution is found quickly and high rewards in later updates if a more stable, faster-converging solution is achieved.

\paragraph{K-Net Reward}
The reward function for NearestK-Net is defined as:
\begin{equation}
\mathcal{R}_{K} = \alpha_K \cdot \frac{1}{t} + \beta_K \cdot \frac{1}{c(\xi)} +\gamma_K \cdot \#\xi,
\label{rewardK}
\end{equation}
specially, \( \#\xi \) denotes the number of states in a solution path.
Since neighbor selection influences path efficiency, this function encourages more direct paths with fewer intermediate states, leading to higher efficiency and better trajectory planning.
This work also incorporates Prioritized Experience Replay (PER)~\cite{schaul2016prioritizedexperiencereplay} to address the sparse rewards problem.
%




\subsubsection{Actor and Critic Network Training}
The objective of the actor network is to maximize the expected Q-value:
\begin{equation}
\bm{W}_{\theta}^{*} =\argmax_{\bm{W}_{\theta}} \mathbb{E} \left[ Q_{\phi}(\mathbf{s}, A_{\theta}(\mathbf{s})) \right].
\label{actor_loss_expectation}
\end{equation}
This is transformed into a loss function for training:
\begin{equation}
\mathcal{L}(\bm{W}_{\theta}) = -\frac{1}{m} \sum_{t} Q_{\phi}(\mathbf{s}, A_{\theta}(\mathbf{s}; \bm{W}_{\theta})).
\label{actorloss}
\end{equation}
where: \( \bm{W}_{\theta} \) represents the actor network parameters, \( m \) is the batch size used during training.

For the critic network, the goal is to minimize the Temporal Difference (TD) error:
\begin{equation}
\mathcal{L}(\bm{W}_{\phi}^{*}) = \argmin_{\bm{W}_{\phi}} \mathbb{E} \left[ \delta^2 \right] 
\label{Critic_loss}
\end{equation}
where \( \delta = Q_{\phi'}(\mathbf{s}, \mathbf{a}) - Q_{\phi}(\mathbf{s}, \mathbf{a}) \) is the TD error.

For mini-batch training, the loss function becomes:
\begin{equation}
\mathcal{L}(\bm{W}_{\theta}) = -\frac{1}{m} \sum_{t} Q_{\phi}(s_t, A_{\theta}(s_t; \bm{W}_{\theta})).
\label{criticloss}
\end{equation}

\subsubsection{Network Update}
To stabilize training, soft updates are applied to the target networks:
\begin{align}
    \bm{W}_{\phi}^{'} &\gets \tau \bm{W}_{\theta} + (1 - \tau) \bm{W}_{\phi}^{'}, \\
    \bm{W}_{\phi}^{'} &\gets \tau \bm{W}_{\phi} + (1 - \tau) \bm{W}_{\phi}^{'}, 
\end{align}
where \( \bm{W}_{\theta}^{'} \) and \( \bm{W}_{\phi}^{'} \) are target network parameters, \( \tau \ll 1 \) is a small update factor ensuring stability.

This gradual update strategy prevents sudden oscillations in policy updates and improves learning efficiency.

Given that the \textit{Fuzzy-DDPG}'s output is deterministic, this study pre-maps the neural network outputs into a tensor (Fig.~\ref{fig:tensor}), where the three tensor dimensions correspond to the three input variables. Each unique three-dimensional coordinate directly determines a specific output value. In C++, querying this tensor has a time complexity of $\bm{O}(1)$, achieving microsecond-level access speed.

The overall process is as follows: during sampling and neighbor node expansion, the current map information is retrieved; based on this information, a lookup is performed in the tensor; the corresponding action is obtained, and the path planning algorithm continues execution.

\section{Experimental Results}\label{sec:experi}

LIT* was evaluated against several existing algorithms, including different versions of RRT-Connect, Informed RRT*, BIT*, AIT*, EIT*, and FIT* from the Open Motion Planning Library (OMPL)~\cite{sucan2012open}. Tests were conducted in simulated environments ranging from $\mathbb{R}^4$ to $\mathbb{R}^{16}$ and simulation manipulation scenarios using an Intel i7 3.90 GHz processor with 32GB of LPDDR3 4800 MHz memory. The main goal was to minimize the median initial path length ($c^\textit{med}_\textit{init}$) over 100 runs. For all planners, the RGG constant $\eta$ was set to 1.1, and the rewire factor to 1.001. RRT-based algorithms used a 5\% goal bias, with maximum edge lengths adjusted for space dimensionality. This work also tests the algorithm in the Open Robotics Automation Virtual Environment (OpenRAVE~\cite{Diankov2008OpenRAVEAP}, Fig.~\ref{fig: simulation} (c) and (d)) by a dual-Barrett Whole-Arm Manipulator (dual-WAM-$\mathbb{R}^{14}$). The learning-based mechanism optimizes the convergence speed and solution quality.

\begin{table}[t]
\caption{Benchmarks evaluation comparison (Fig.~\ref{fig: result})}
\centering
\resizebox{0.485\textwidth}{!}{
\begin{tabular}{p{1.5cm}ccccccc}
 \toprule
 & \multicolumn{3}{c}{${\text{Flexible Informed Trees~\cite{Zhang2024adaptive}}}$} & \multicolumn{3}{c}{$\textcolor{purple}{\text{Learning-based                                  Informed Trees}}$} &\multirow{2}*{\large$t^\textit{med}_\textit{init}\color{purple}\Uparrow$ (\%)}\\
 \cmidrule(lr){2-4} \cmidrule(lr){5-7}
    &$t^\textit{med}_\textit{init}$ &$c^\textit{med}_\textit{init}$ &$c^\textit{med}_\textit{final}$ &$t^\textit{med}_\textit{init}$ &$c^\textit{med}_\textit{init}$ &$c^\textit{med}_\textit{final}$ \\
 \midrule

    $\text{RR}-\mathbb{R}^4$   &0.0626   &2.3295   &1.9951 &0.0597 &\textcolor{purple}{1.0621} &1.8383 &4.63  \\
    $\text{RR}-\mathbb{R}^8$   &0.1247   &3.7650   &3.1704 &0.1114 &\textcolor{purple}{3.1188} &2.8034  &{10.67}  \\
    $\text{RR}-\mathbb{R}^{16}$   
    &{0.2778} &{6.2709} &{4.2460} &{0.2385} &\textcolor{purple}{4.9677} &{3.9552}  &{14.15}\\
\midrule    
    $\text{NP}-\mathbb{R}^4$    &0.0084 &1.3681 &0.5395 &\textcolor{purple}{0.0060} &1.1506 &0.5096  &\textcolor{purple}{28.6}\\
    $\text{NP}-\mathbb{R}^8$  &0.0291 &2.7310  &2.0761 &\textcolor{purple}{0.0217} &2.5802 &\textcolor{purple}{1.9441} &\textcolor{purple}{25.43}  \\
    $\text{NP}-\mathbb{R}^{16}$   &0.0383   &4.7280   &3.8258 &\textcolor{purple}{0.0262} &\textcolor{purple}{4.0591} &\textcolor{purple}{2.9497}  &\textcolor{purple}{\textbf{31.59}}  \\
\bottomrule
\end{tabular}} \label{tab:benchmark}
\vspace{-1.7em} 
\end{table}

As observed in Table \ref{tab:benchmark}, there's a median cost improvement across varied benchmark scenarios, correlating with dimensionality. In the case of the NP-$\mathbb{R}^{16}$ scenario, the initial median solution cost exhibits a reduction of up to 31.59\%.
\section{Discussion \string& Conclusion}

This paper introduces Learning-based Informed Trees (LIT*), a planner using \textit{Fuzzy-DDPG} to determine batchsize and nearest neighbors. It also encodes the invalid states as input of the \textit{Fuzzy-DDPG} framework, thereby self-adjusting according to the valid/invalid ratio. 
Since multiple neighbor expansion steps occur in each epoch, and the motion planning algorithm is implemented in C++ while Fuzzy-DDPG runs in Python, the communication between the two can be significant. This work facilitates data exchange by a shared file, which may introduce latency. 
Future work could focus on improving communication efficiency to reduce training time. LIT* demonstrated its adaptability by achieving short path lengths and quickly generating solutions.

In conclusion, LIT* leverages valid and invalid states to dynamically determine batchsize and the number of neighbors, ensuring adaptive and efficient motion planning.

\bibliographystyle{IEEEtran}
\bibliography{references}

\begin{thebibliography}{10}
\providecommand{\url}[1]{#1}
\csname url@samestyle\endcsname
\providecommand{\newblock}{\relax}
\providecommand{\bibinfo}[2]{#2}
\providecommand{\BIBentrySTDinterwordspacing}{\spaceskip=0pt\relax}
\providecommand{\BIBentryALTinterwordstretchfactor}{4}
\providecommand{\BIBentryALTinterwordspacing}{\spaceskip=\fontdimen2\font plus
\BIBentryALTinterwordstretchfactor\fontdimen3\font minus \fontdimen4\font\relax}
\providecommand{\BIBforeignlanguage}[2]{{%
\expandafter\ifx\csname l@#1\endcsname\relax
\typeout{** WARNING: IEEEtran.bst: No hyphenation pattern has been}%
\typeout{** loaded for the language `#1'. Using the pattern for}%
\typeout{** the default language instead.}%
\else
\language=\csname l@#1\endcsname
\fi
#2}}
\providecommand{\BIBdecl}{\relax}
\BIBdecl

\bibitem{zhang2024review}
L.~Zhang, K.~Cai, Z.~Sun, Z.~Bing, C.~Wang, L.~Figueredo, S.~Haddadin, and A.~Knoll, ``Motion planning for robotics: A review for sampling-based planners,'' \emph{Biomimetic Intelligence and Robotics}, vol.~5, no.~1, p. 100207.

\bibitem{dijkstra1959note}
E.~Dijkstra, ``A note on two problems in connexion with graphs,'' \emph{Numerische Mathematik}, vol.~1, no.~1, pp. 269--271, 1959.

\bibitem{hart1968formal}
P.~E. Hart, N.~J. Nilsson, and B.~Raphael, ``A formal basis for the heuristic determination of minimum cost paths,'' \emph{IEEE Transactions on Systems Science and Cybernetics}, vol.~4, no.~2, pp. 100--107, 1968.

\bibitem{LaValle1998RRT}
S.~M. LaValle, ``Rapidly-exploring random trees: A new tool for path planning,'' Iowa State University, Tech. Rep. 98-11, October 1998.

\bibitem{Kavraki1996PRM}
L.~E. Kavraki, P.~Svestka, J.-C. Latombe, and M.~H. Overmars, ``Probabilistic roadmaps for path planning in high-dimensional configuration spaces,'' \emph{IEEE Transactions on Robotics and Automation}, vol.~12, no.~4, pp. 566--580, 1996.

\bibitem{Karaman2011RRTstar}
S.~Karaman and E.~Frazzoli, ``Sampling-based algorithms for optimal motion planning,'' \emph{The International Journal of Robotics Research}, vol.~30, no.~7, pp. 846--894, 2011.

\bibitem{gammell2014informed}
J.~D. Gammell, S.~S. Srinivasa, and T.~D. Barfoot, ``Informed {RRT}*: Optimal sampling-based path planning focused via direct sampling of an admissible ellipsoidal heuristic,'' in \emph{2014 IEEE/RSJ international conference on intelligent robots and systems}.\hskip 1em plus 0.5em minus 0.4em\relax IEEE, 2014.

\bibitem{fmt2015}
L.~Janson, E.~Schmerling, A.~Clark, and M.~Pavone, ``Fast marching tree: A fast marching sampling-based method for optimal motion planning in many dimensions,'' \emph{The International Journal of Robotics Research}, vol.~34, no.~7, pp. 883--921, 2015.

\bibitem{gammell2020batch}
J.~D. Gammell, T.~D. Barfoot, and S.~S. Srinivasa, ``Batch informed trees ({BIT}*): Informed asymptotically optimal anytime search,'' \emph{The International Journal of Robotics Research}, vol.~39, no.~5, 2020.

\bibitem{strub2022adaptively}
M.~P. Strub and J.~D. Gammell, ``Adaptively informed trees ({AIT}*) and effort informed trees ({EIT*}): Asymmetric bidirectional sampling-based path planning,'' \emph{The International Journal of Robotics Research}, vol.~41, no.~4, pp. 390--417, 2022.

\bibitem{zhang25ral}
L.~Zhang, Y.~Ling, Z.~Bing, F.~Wu, S.~Haddadin, and A.~Knoll, ``Tree-based grafting approach for bidirectional motion planning with local subsets optimization,'' \emph{IEEE Robotics and Automation Letters}, vol.~10, no.~6, pp. 5815--5822, 2025.

\bibitem{ZHANG2025git}
L.~Zhang, K.~Cai, Z.~Bing, C.~Wang, and A.~Knoll, ``Genetic informed trees ({GIT*}): Path planning via reinforced genetic programming heuristics,'' \emph{Biomimetic Intelligence and Robotics}, vol.~5, no.~3, p. 100237, 2025.

\bibitem{Qureshi2020MPNet}
A.~H. Qureshi, Y.~Miao, A.~Simeonov, and M.~C. Yip, ``Motion planning networks: Bridging the gap between learning-based and classical motion planners,'' \emph{IEEE Transactions on Robotics}, vol.~37, no.~1, pp. 48--66, 2021.

\bibitem{chen2020learningplanhighdimensions}
\BIBentryALTinterwordspacing
B.~Chen, B.~Dai, Q.~Lin, G.~Ye, H.~Liu, and L.~Song, ``Learning to plan in high dimensions via neural exploration-exploitation trees,'' 2020. [Online]. Available: \url{https://arxiv.org/abs/1903.00070}
\BIBentrySTDinterwordspacing

\bibitem{NeuralInformedRRT}
Z.~Huang, H.~Chen, J.~Pohovey, and K.~Driggs-Campbell, ``Neural informed {RRT}*: Learning-based path planning with point cloud state representations under admissible ellipsoidal constraints,'' in \emph{2024 IEEE International Conference on Robotics and Automation (ICRA)}, 2024, pp. 8742--8748.

\bibitem{DeepFuzzyRobotSpeed}
F.~Wu, W.~Tang, Y.~Zhou, S.-W. Lin, Y.~Liu, and Z.~Ding, ``Continuous motion planning for mobile robots using fuzzy deep reinforcement learning,'' in \emph{2024 WRC Symposium on Advanced Robotics and Automation (WRC SARA)}.\hskip 1em plus 0.5em minus 0.4em\relax IEEE, 2024, pp. 15--21.

\bibitem{zhang2025TASE}
L.~Zhang, K.~Cai, Y.~Zhang, Z.~Bing, C.~Wang, F.~Wu, S.~Haddadin, and A.~Knoll, ``Estimated informed anytime search for sampling-based planning via adaptive sampler,'' \emph{IEEE Transactions on Automation Science and Engineering}, vol.~22, pp. 18\,580--18\,593, 2025.

\bibitem{zhang2025apt}
L.~Zhang, S.~Wang, K.~Cai, Z.~Bing, F.~Wu, C.~Wang, S.~Haddadin, and A.~Knoll, ``{APT*}: Asymptotically optimal motion planning via adaptively prolated elliptical r-nearest neighbors,'' \emph{IEEE Robotics and Automation Letters}, vol.~10, no.~10, pp. 10\,242--10\,249, 2025.

\bibitem{Zhang2024adaptive}
L.~Zhang, Z.~Bing, K.~Chen, L.~Chen, K.~Cai, Y.~Zhang, F.~Wu, P.~Krumbholz, Z.~Yuan, S.~Haddadin, and A.~Knoll, ``Flexible informed trees ({FIT}*): Adaptive batch-size approach in informed sampling-based path planning,'' \emph{2024 IEEE/RSJ International Conference on Intelligent Robots and Systems (IROS)}, pp. 3146--3152, 2024.

\bibitem{Zhang2024Elliptical}
L.~Zhang, Z.~Bing, Y.~Zhang, K.~Cai, L.~Chen, F.~Wu, S.~Haddadin, and A.~Knoll, ``Elliptical k-nearest neighbors: Path optimization via coulomb's law and invalid vertices in c-space obstacles,'' \emph{2024 IEEE/RSJ International Conference on Intelligent Robots and Systems (IROS)}, pp. 12\,032--12\,039, 2024.

\bibitem{andoni2009nearest}
A.~Andoni, ``Nearest neighbor search: the old, the new, and the impossible,'' Ph.D. dissertation, Massachusetts Institute of Technology, 2009.

\bibitem{FuzzyMachineLearning}
J.~Lu, G.~Ma, and G.~Zhang, ``Fuzzy machine learning: A comprehensive framework and systematic review,'' \emph{IEEE Transactions on Fuzzy Systems}, vol.~32, no.~7, pp. 3861--3878, 2024.

\bibitem{karaman2011sampling}
S.~Karaman and E.~Frazzoli, ``Sampling-based algorithms for optimal motion planning,'' \emph{The international journal of robotics research}, vol.~30, no.~7, pp. 846--894, 2011.

\bibitem{gammell2022planner}
J.~D. Gammell, M.~P. Strub, and V.~N. Hartmann, ``Planner developer tools (pdt): Reproducible experiments and statistical analysis for developing and testing motion planners,'' in \emph{Proc. Workshop EMPP, IEEE/RSJ IROS}, 2022.

\bibitem{Diankov2008OpenRAVEAP}
R.~Diankov and J.~J. Kuffner, ``Openrave: A planning architecture for autonomous robotics,'' in \emph{Carnegie Mellon University}, 2008.

\bibitem{softmaxNeuralNetwork}
Y.~Cui, D.~Wu, and Y.~Xu, ``Curse of dimensionality for tsk fuzzy neural networks: Explanation and solutions,'' in \emph{2021 International Joint Conference on Neural Networks (IJCNN)}, 2021, pp. 1--8.

\bibitem{TSKDefuzzy}
Y.~Wang, H.~Liu, W.~Jia, S.~Guan, X.~Liu, and X.~Duan, ``Deep fuzzy rule-based classification system with improved wang–mendel method,'' \emph{IEEE Transactions on Fuzzy Systems}, vol.~30, no.~8, pp. 2957--2970, 2022.

\bibitem{schaul2016prioritizedexperiencereplay}
\BIBentryALTinterwordspacing
T.~Schaul, J.~Quan, I.~Antonoglou, and D.~Silver, ``Prioritized experience replay,'' 2016. [Online]. Available: \url{https://arxiv.org/abs/1511.05952}
\BIBentrySTDinterwordspacing

\bibitem{sucan2012open}
I.~A. Sucan, M.~Moll, and L.~E. Kavraki, ``The open motion planning library,'' \emph{IEEE Robotics \& Automation Magazine}, vol.~19, no.~4, pp. 72--82, 2012.

\end{thebibliography}
\balance

\end{document}